%% file: main.tex
\documentclass[runningheads]{llncs}

\usepackage[T1]{fontenc}
\usepackage{graphicx}

\usepackage{hyperref}
\usepackage[table]{xcolor}
\usepackage{color}
\usepackage[skins]{tcolorbox}
\usepackage{booktabs}
\usepackage{tabularx}
\usepackage{subcaption}
\usepackage{multirow}
\usepackage{makecell}
\usepackage{float}
\usepackage[linesnumbered,ruled,vlined]{algorithm2e}
\newfloat{Algorithm}{H}{alg}

\usepackage[symbol*]{footmisc}

\usepackage[linesnumbered,ruled,vlined]{algorithm2e}
\DontPrintSemicolon       
\SetAlgoNlRelativeSize{-1} 
\SetAlgoNlRelativeSize{0} 
\SetAlgoNlRelativeSize{-1} 

\begin{document}

\title{Robustness of Misinformation Classification Systems to Adversarial Examples Through BeamAttack}

\titlerunning{Robustness of Misinformation Classification through BeamAttack}

\author{Arnisa Fazla\,\thanks{These authors contributed equally. }\protect\inst{1} \and
Lucas Krauter\,\protect\footnotemark[1]\protect\inst{2} \and
\\ David Guzman Piedrahita\,\protect\footnotemark[1]\protect\inst{2} \and
Andrianos Michail\,\protect\inst{3}}

\institute{Department of Medical Informatics, Amsterdam UMC \\
\email{a.fazla@amsterdamumc.nl} \\
\and Department of Informatics, University of Zurich \\
\email{\{lucassteffen.krauter,david.guzmanpiedrahita\}@uzh.ch} \\
\and Department of Computational Linguistics, University of Zurich \\
\email{andrianos.michail@cl.uzh.ch}}


\authorrunning{A. Fazla et al.}

\maketitle          

\renewcommand\thefootnote{\arabic{footnote}}
\setcounter{footnote}{0}

\begin{abstract}
We extend BeamAttack, an adversarial attack algorithm designed to evaluate the robustness of text classification systems through word-level modifications guided by beam search. Our extensions include support for word deletions and the option to skip substitutions, enabling the discovery of minimal modifications that alter model predictions. We also integrate LIME to better prioritize word replacements. Evaluated across multiple datasets and victim models (BiLSTM, BERT, and adversarially trained RoBERTa) within the BODEGA framework, our approach achieves over a 99\% attack success rate while preserving the semantic and lexical similarity of the original texts. Through both quantitative and qualitative analysis, we highlight BeamAttack's effectiveness and its limitations. Our implementation is available at \url{https://github.com/LucK1Y/BeamAttack}.
\keywords{Model Robustness \and Adversarial Attack \and Beam search \and Masked Language Models.}
\end{abstract}

\section{Introduction}

\begin{figure}[b]
    \centering
    \includegraphics[width=\linewidth]{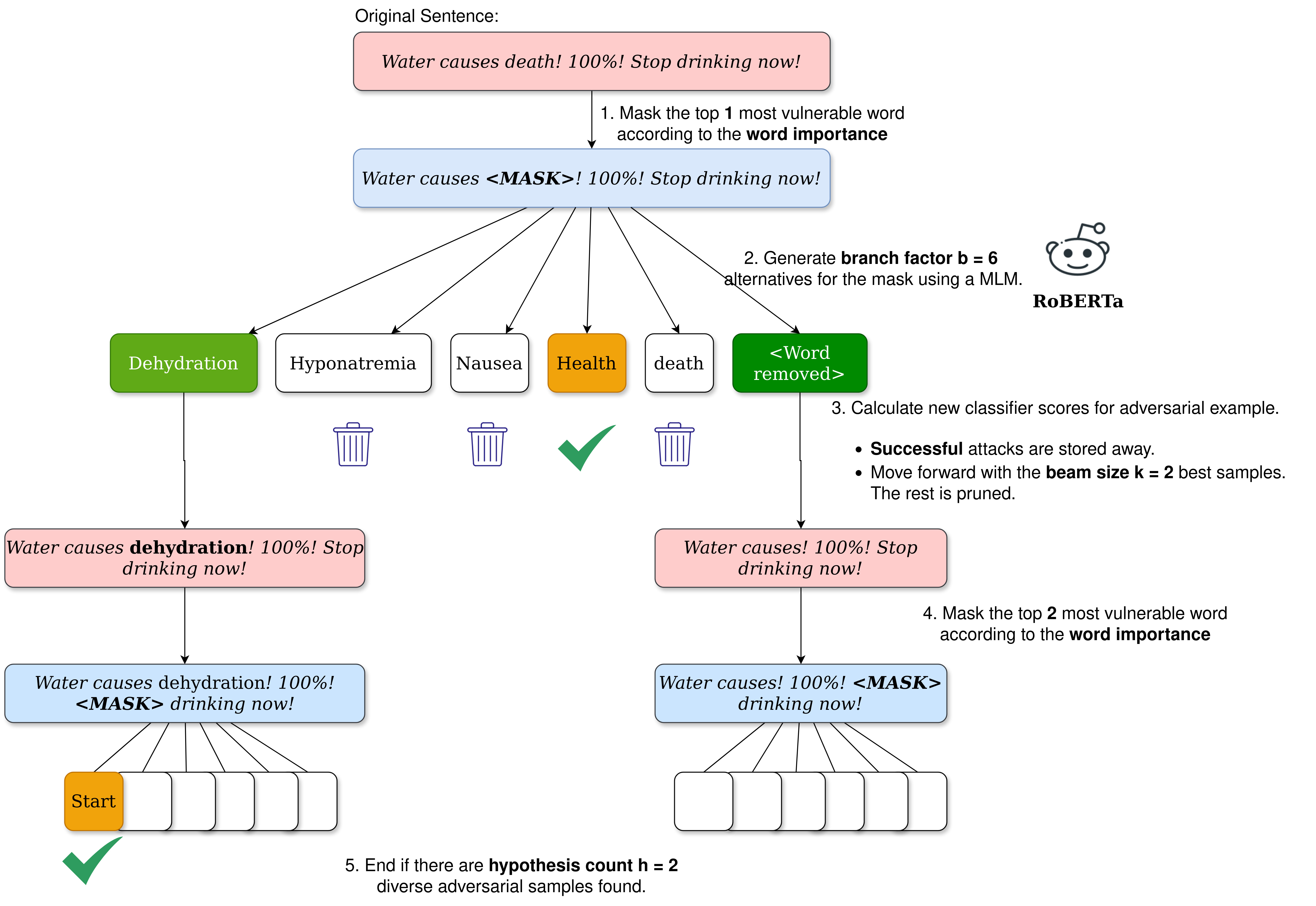}
    \caption{Diagram depicting the series of actions in our BeamAttack algorithm, where beam size \( k \) = 2, branching factor \( b \) = 6, hypothesis count \( h \) = 2. }
    \label{fig:beam-attack}
\end{figure}

Social media platforms increasingly rely on machine learning algorithms for content filtering to identify misleading, harmful, or illegal content. Despite advancements, these models remain vulnerable to adversarial attacks, wherein the input text is manipulated to deceive the model. An adversarial sample in the text classification domain is generated from the original sample by applying the smallest possible change that leads to a misclassification by the target model, known as the victim classifier. The goal is to alter the sample just enough to trick the classifier into making an incorrect prediction, while preserving the original meaning for human interpretation \cite{advAttacK_Survey}. Adversarial training (continued training of the models with adversarial samples to improve the resilience against such attacks) can make the victim models more robust, but novel attack methods can succeed even when tested on adversarially trained models. As adversaries continue to devise novel attack strategies that exploit vulnerabilities models were not trained to handle, we need to develop new attack techniques for assessing the robustness of text classification systems and finding potential weaknesses. These weaknesses can then be mitigated through adversarial training with samples from these novel techniques.

Previous adversarial attacks on text classification systems have explored character-level modifications such as DeepWordBug \cite{replace-with-oov-attack}, word-level modifications such as TextBugger \cite{textbugger-word-char-lvl}, BERT-ATTACK \cite{bert-attack} and BeamAttack \cite{10.1007/978-3-031-33377-4_35}, and sentence level paraphrasing of the input text such as SEAs \cite{ribeiro-etal-2018-semantically}. While these approaches have shown promise, there remains a need for comprehensive evaluation frameworks that assess attack effectiveness across diverse datasets and victim models, particularly those that have undergone adversarial training.

BeamAttack \cite{10.1007/978-3-031-33377-4_35} runs a beam search with the heuristic of maximizing the victim model's output probability for the target class while minimizing the number of word modifications. It represents an effective grey-box approach that balances attack success with semantic preservation. Experiments show that BeamAttack achieves trade-off results compared with previous works, with adversarial examples crafted by BeamAttack demonstrating high semantic similarity, low perturbation, and good transferability.

However, BeamAttack relies on logit-based word importance ranking to determine which words to target for substitution. LIME (Local Interpretable Model-agnostic Explanations) \cite{lime} offers a more precise alternative by training local linear models to estimate how each token influences a specific prediction. Unlike global explanation methods, LIME's local approach is ideal for adversarial attacks since we need to understand token importance for individual samples rather than model behavior overall. By prioritizing words with the highest LIME-derived importance scores, we hypothesize that attacks can achieve misclassification with fewer modifications, focusing perturbations on tokens that most directly influence the model's decision for that specific input.

We present an extended version of BeamAttack, evaluated within the BODEGA framework \cite{bodega-paper}, which provides standardized metrics and datasets for benchmarking adversarial attacks on text classifiers. Our implementation was originally developed as part of our submission \cite{piedrahita2024texttrojaners} to Task 6 of the CheckThat! Lab at CLEF 2024 \cite{clef-checkthat:2024:task6}. Like the original BeamAttack, our approach uses beam search to identify optimal word-level modifications and generate multiple adversarial hypotheses capable of flipping the model’s prediction. We rely on a Masked Language Model (MLM), specifically RoBERTa \cite{roberta}, to produce contextually appropriate substitution candidates.

Although our core algorithmic approach is similar to BeamAttack, our work makes several unique contributions. We have developed extensive evaluation frameworks and novel analysis methodologies that provide insights into the effectiveness of beam search-based adversarial attacks in various contexts. Our comprehensive analysis aims to deepen the understanding of adversarial attack mechanisms and identify opportunities to improve BeamAttack. Our key contributions are the following:

\begin{itemize}
    \item \textbf{LIME for Word Importance Ranking:} Beyond the standard logit-based calculation of word importance, which is common in word-level adversarial attacks \cite{textbugger-word-char-lvl,bert-attack,10.1007/978-3-031-33377-4_35}, we introduce and evaluate LIME-based \cite{lime} word importance ranking as an alternative strategy for identifying vulnerable words in an input text. We demonstrate that, for certain specific tasks (\textit{HN} and \textit{FC}) and transformer based encoders, using LIME to determine which words to replace achieves better results than the logit-based importance method, albeit with a significantly higher number of queries.
    \item \textbf{Token Removal:}  We extend the original BeamAttack by incorporating token removal as an additional modification strategy, examining its impact on attack success rates and human assessment of semantic preservation of the original text.
    \item \textbf{Cross-Model Evaluation:} We provide a comprehensive analysis of BeamAttack across diverse victim models (BiLSTM, BERT, and adversarially trained RoBERTa) and datasets. This analysis offers insights into attack transferability and model-specific vulnerabilities.
    \item \textbf{Linguistic and Quality Analysis:} We conduct both automatic and manual evaluations of successful adversarial samples, focusing on word substitution rates (WSR) and part-of-speech (POS) tag changes. By analyzing POS transition patterns across tasks and victim models, we gain insights into how linguistic structures are manipulated during attacks, how these affect perceived quality, and what they reveal about task difficulty and model robustness—informing future defense strategies.
\end{itemize}

\section{Related Work}
\label{sec:background}

Adversarial text attacks can be categorized by their granularity of modification. Character-level methods change individual characters, such as DeepWordBug \cite{replace-with-oov-attack}, which replaces characters to make words unrecognizable while remaining human-readable. Word-level methods modify text by substituting, adding, or removing words, exemplified by approaches that replace words with synonyms or out-of-vocabulary terms \cite{adversarial-attack-replace-with-grammatical-synonym,bert-attack}. Sentence-level methods paraphrase entire sentences, such as SCPN \cite{iyyer2018adversarial}, which generates adversarial examples through whole-text paraphrasing. Most recent work follows a search-based paradigm that ranks words by importance and searches for effective replacements using strategies like synonym substitution, masked language models \cite{BAE,bert-attack}, or evolutionary algorithms \cite{adversarial-text-genetic-optimization}.

Attacks are further categorized by objective and access level. Targeted attacks aim to mislead models into specific incorrect classes, while untargeted attacks seek any misclassification. Regarding access, white-box attacks leverage full model access for gradient-based optimization \cite{white-box-adversarial-attack-text-classification}, black-box attacks use only predictions with surrogate models \cite{query-efficient-importances}, and grey-box attacks access both predictions and probability distributions \cite{bodega-paper,bert-attack}.

Our method follows the established search-based research paradigm, operating as a grey-box, word-level, untargeted attack that leverages probability distributions to guide adversarial example generation.

Our method follows the established search-based paradigm as a grey-box, word-level, untargeted attack. We evaluate using the BODEGA score \cite{bodega-paper}, which is a composite metric incorporating three components: the confusion score, which indicates whether the victim classifier’s decision has changed and the attack was successful; the semantic score, which measures the similarity between the original and adversarial examples using BLEURT \cite{bleurt}; and the character score, which calculates the Levenshtein distance \cite{levenshtein1966binary} as a character-level similarity measure. All three metrics range from 0 to 1, with higher values indicating greater success or similarity. Evaluation is conducted within the InCrediblAE framework from CheckThat! lab at CLEF 2024 \cite{clef-checkthat-2024,clef-checkthat:2024:task6}.

\section{Methodology}
\label{sec:methodology}

We build upon BeamAttack \cite{10.1007/978-3-031-33377-4_35}, extending it with additional word importance methods and replacement strategies. Our approach reframes word-level attacks as search problems, where the search space is determined by the number of words to be replaced and valid replacements for each word.\footnote{The pseudocode for the algorithm can be found in Appendix \ref{sec:algo}.}

Most existing techniques employ greedy search, replacing words sequentially by importance order until the model's prediction changes \cite{adversarial-attack-replace-with-wordnet,adversarial-attack-replace-with-grammatical-synonym,query-efficient-importances,bert-attack}, which can lead to sub-optimal adversarial samples requiring unnecessary word changes.

We adapt beam search for adversarial text generation, where it explores multiple solutions concurrently in a tree structure. Each node represents a partial adversarial text and branches signify word substitutions or removals. Beam search maintains \( k \) most promising partial solutions at each step, balancing exploration with computational efficiency.

Our implementation operates by iteratively expanding promising nodes until reaching final adversarial examples. The root node represents the original input text, and each subsequent level adds one word modification. At each depth, we evaluate \( k \) × \( b \) nodes (\( k \) retained nodes from the previous level, each expanded with \( b \) word replacement candidates) and prune back to the \( k \) most promising solutions. Node selection prioritizes effectiveness in reducing the original class probability while increasing the target class probability.

The search continues until generating \( h \) successful adversarial samples that flip the model's classification. The final adversarial example is selected as the hypothesis with highest semantic similarity to the original text, measured by BLEURT scores \cite{bleurt}. This approach generates adversarial samples with fewer modifications while maintaining semantic coherence.

Our extended BeamAttack algorithm consists of two core components, both integrated with the beam search procedure: a method for determining the order in which words are modified (word importance ranking) and a strategy for how those words are replaced (word replacement strategy). An overview of the full algorithm is provided in Figure~\ref{fig:beam-attack}. We enhance the original BeamAttack by incorporating LIME to compute word importance scores and by introducing word deletion as an additional replacement option.

\subsection{Word Replacement Order}
We evaluate two techniques for identifying the most important words: LIME \cite{lime} and logit-based importance scores \cite{bert-attack}.

LIME provides local explanations by perturbing input data and observing output changes. For each word, it generates perturbed sentence versions with word replacements, then fits an interpretable model to estimate word importance based on output variations.

Logit-based importance measures each word's influence on the model's output logits by removing the word and observing the resulting change in the logits. For input sentence \( S = [w_0, \ldots, w_i, \ldots] \) and logit output \( o_y(S) \) for the correct label \( y \). The importance score \( I_{w_i} \) for each word \( w_i \) is defined as the difference in logits when the word is masked:

\[ I_{w_i} = o_y(S) - o_y(S \setminus w_i), \]

where \( S \setminus w_i \) is the sentence with \( w_i \) replaced by a [MASK] token. This score quantifies how much the presence of \( w_i \) contributes to the model's prediction.

\subsection{Word Replacement Strategy}
For each individual replacement, we enable the option to either keep the word intact or remove it altogether at any depth of the beam search. This added flexibility allows us to preserve the original word if replacements have no impact on changing the probability, thus reducing similarity and edit distances. Conversely, removing the word entirely may be the best strategy to confuse the model in more challenging cases, even though it might introduce grammar errors and inconsistencies into the adversarial sample. Following BeamAttack \cite{10.1007/978-3-031-33377-4_35}, we use a Masked Language Model (MLM) to leverage the model’s pretraining to ensure the suggestions fit the context fluently.

For each depth of beam search, we evaluate the top \( b \) (branching factor) highest likelihood word replacements as suggested by the masked language model. This strategy ensures that the potential replacements not only fit well within the context of the sentence but also maintain the overall semantic coherence of the text.

\subsection{Hyperparameters}

For each dataset and victim model combination, we determined distinct hyperparameters for our BeamAttack approach through an initial exploration on the development sets (Appendix \ref{sec:dataset_statistics}). While this approach may not yield globally optimal combinations, it balanced effectiveness with computational constraints. Complete hyperparameter configurations and resource usage details are provided in Appendices \ref{sec:hyperparameter_search} and \ref{sec:resources}, respectively.

To reduce query overhead, we also lowered LIME’s sample count from 5000 to 500. Initial observations showed this had minimal impact on attack quality for shorter texts, though it may affect longer samples. We leave further tuning of LIME’s kernel parameters for future work.

\section{Results}
\label{sec:results}

\input{results_all}

We report the automatic evaluation results of BeamAttack in Table \ref{tab:results_combined}, compared to the two strongest methods from previous work, BERT-ATTACK \cite{bert-attack} and DeepWordBug \cite{replace-with-oov-attack}, as well as OpenFact \cite{lewoniewski2024openfact}, the top-performing system submitted to the CheckThat! 2024 Lab Task 6. The complete results, including all baseline methods and all submissions to the shared task, can be found in the shared task overview paper \cite{clef-checkthat:2024:task6}.

On automatic evaluation, our algorithm outperforms all other algorithms on BERT and  BiLSTM victims, demonstrating the superior effectiveness of BeamAttack. While other algorithms may generate adversarial samples with better character or automatically measured semantic preservation score in specific tasks, they fail to achieve this consistently across the entire dataset. Moreover, our confusion scores (success rates) are always superior to those of other techniques. 

Nevertheless, our algorithm needs many victim queries to achieve this goal. For example, in the task \textit{RD} on the BERT classifier, our BeamAttack takes roughly double the number of queries compared to the Genetic algorithm (The details on number of queries to the victim model are reported in Appendix Table \ref{tab:num_queries}). This highlights a trade-off between the quality of the generated adversarial samples and the computational resources required to achieve them.

\subsection{Manual Evaluation Results}
In contrast to the automatic evaluation results, the human evaluation of semantic preservation revealed a different trend. Our team, TextTrojaners, received the lowest scores in this round, with only 7\% of our adversarial samples categorized as \texttt{Preserving the Original Meaning}, while 63\% were labeled as \texttt{Changing the Original Meaning}, and 30\% as \texttt{Nonsensical} \cite{clef-checkthat:2024:task6}. Notably, the manually evaluated subset consisted solely of samples from the \textit{FC} dataset, with 96\% targeting the adversarially trained RoBERTa model. This outcome indicates that our attack strategy, although effective against automated systems, often altered the original semantics.

One possible explanation for the 63\% of samples labeled as semantically changed is the use of masked language models (MLMs) for word replacement. While these replacements are contextually and grammatically plausible, they do not explicitly aim to preserve the meaning of the original token, often introducing substitutions such as antonyms or semantically divergent alternatives. For the 30\% of samples deemed nonsensical, we suspect that our word removal strategy significantly contributes to syntactic breakdowns. To better understand these failure cases, we conduct an in-depth manual analysis of these ungrammatical cases in Section~\ref{subsection:analysis_manual_evaluation}, identifying specific patterns and causes of structural degradation.

\section{Analysis of Output Adversarial Samples and Discussion}
\label{sec:analysis}

We analyze structural differences between original and adversarial samples through POS tag sequences and word substitution ratios. Our analysis covers 6,503 successful adversarial examples (97.16\% success rate) to examine the magnitude and nature of perturbations.

\begin{figure}[h]
    \centering
    \includegraphics[width=1\linewidth]{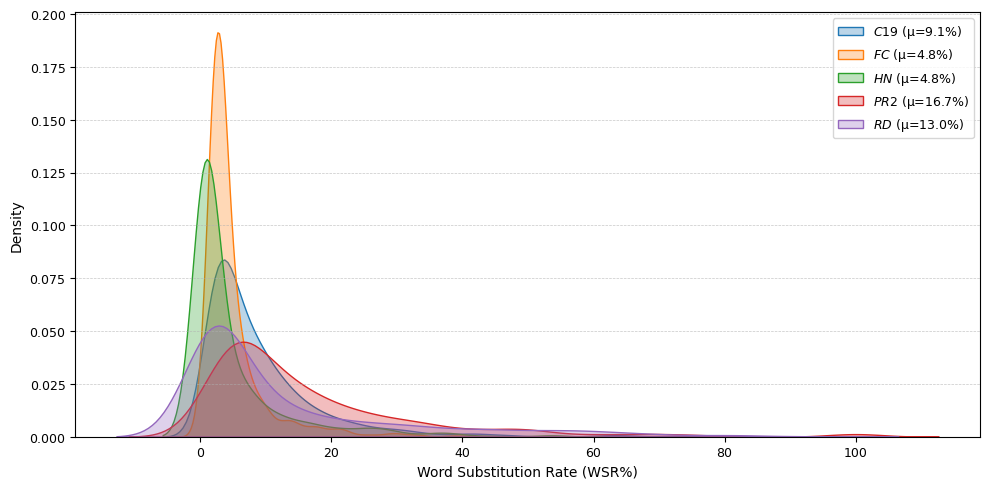}
    \caption{Kernel Density Estimation (KDE) plot showing the density distribution of Word Substitution Rate (WSR (\%)) \cite{10.1007/978-3-031-33377-4_35} per adversarial sample, grouped by task and aggregated over all victim models. Mean WSR for each task is indicated in the legend. Simpler tasks such as \textit{FC} and \textit{HN} exhibit both lower mean WSR and smaller standard deviations compared to the other three tasks.}
    \label{fig:word_change_per_task_kde_all_samples}
\end{figure}

Figure~\ref{fig:word_change_per_task_kde_all_samples} presents the distribution of word-level edit ratios required to fool each victim model, serving as a proxy for task difficulty. Longer texts (\textit{HN}: ~1,972 words) show lower relative substitution rates (~4.8\%) than shorter texts, indicating that adversarial perturbations must be more concentrated in longer documents to achieve success.

\begin{figure}[h]
    \centering
    \begin{subfigure}[b]{0.42\linewidth}
        \includegraphics[width=\linewidth]{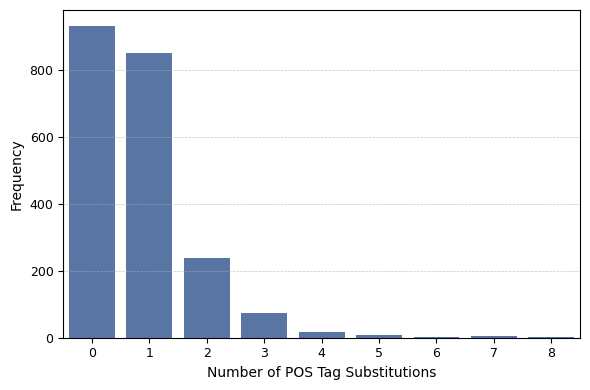}
        \caption{Distribution of number of POS tag substitutions for successful samples with a single word-substitution, where the number of POS tags remained unchanged, aggregated across all tasks and victim models.}
        \label{fig:pos_change_all_samples}
    \end{subfigure}
    \hfill
    \begin{subfigure}[b]{0.54\linewidth}
        \includegraphics[width=\linewidth]{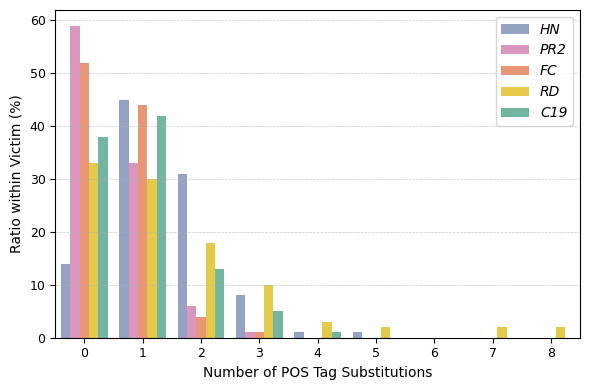}
        \caption{Normalized distribution of number of POS tag substitutions for successful samples with a single word substitution, where the number of POS tags remained unchanged. Frequencies are normalized within each task and aggregated over all victim models. }
        \label{fig:pos_change_per_task_all_samples}
    \end{subfigure}
    \caption{Histograms of POS tag changes across and within tasks.}
    \label{fig:pos_change_combined}
\end{figure}

\begin{figure}[h]
    \centering
    \includegraphics[width=0.9\linewidth]{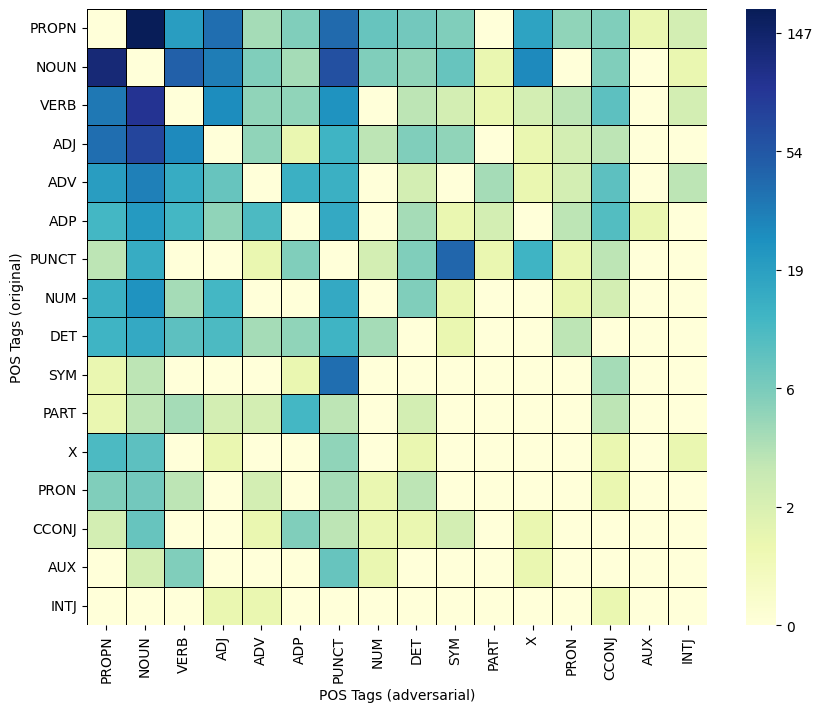}
    \caption{POS transition matrix for the successful samples where the length of POS tags remained the same, and only one POS tag was substituted to another, aggregated over all tasks and victim models.}
    \label{fig:pos_transition_matrix_all_samples}
\end{figure}

We label original and adversarial texts using the \texttt{flair/pos-english} model\footnote{Accessible at: \url{https://huggingface.co/flair/pos-english}}, selected for its contextual understanding and fine-grained tagging capabilities \cite{akbik-etal-2019-flair}. Our analysis focuses on 2,130 single-word substitutions that preserve POS sequence length (90.48\% of single substitutions), as deviations likely indicate tagging noise.\footnote{Examples of outliers where a single-word substitution led to a five-token POS tag change in one case, and an eight-token change in another, can be found in Appendix~\ref{appendix:pos_tagging_examples}.} Among these samples, 933 (43.8\%) show no POS tag changes, while 851 (39.95\%) contain exactly one change, following a long-tailed distribution where most edits are syntactically minimal (Figure \ref{fig:pos_change_all_samples}). This indicates that lexical replacements often preserve syntactic structure.

Task-specific analysis (Figure~\ref{fig:pos_change_per_task_all_samples}) reveals notable differences across datasets. \textit{PR2} samples exhibit the fewest changes, with 58\% remaining unchanged, suggesting that adversarial outputs often retain grammatical consistency. In contrast, \textit{RD} shows a broader distribution, with higher frequencies of 2- and 3-token POS tag substitutions, and a relatively higher occurrence of samples with 7 or 8 changes. This may be attributed to differences in dataset sources: \textit{PR2} (Propaganda Detection) texts are drawn from structured news articles, while \textit{RD} (Rumor Detection) texts originate from informal Twitter threads, which are often unstructured and include links, user mentions, and varied formatting. These characteristics likely make it more challenging for the MLM to substitute tokens while preserving grammatical structure.

Figure~\ref{fig:pos_transition_matrix_all_samples} displays the POS transition matrix from samples with single POS tag substitutions. The matrix appears roughly symmetrical, indicating that similar POS tags can be switched interchangeably. Notable patterns include frequent transitions between PROPN (Proper Nouns) and NOUN, as well as between SYM (Symbols) and PUNC (Punctuation), suggesting these categories are particularly susceptible to adversarial substitution.

\subsection{Manual Evaluation Analysis}
\label{subsection:analysis_manual_evaluation}

We analyze adversarial samples labeled as \texttt{Nonsensical} to identify failure modes. Our analysis reveals two primary substitution categories: (1) grammatically-preserving noun replacements that maintain structure while introducing semantic absurdity, and (2) structure-breaking replacements that alter grammatical integrity.

\begin{figure}[ht]
\centering
\begin{minipage}{0.95\textwidth}
\small
\texttt{Original:} \quad \colorbox{yellow!50}{"Fanaticism} is a monster \colorbox{yellow!50}{that} pretends to \colorbox{yellow!50}{be} the \colorbox{yellow!50}{child of religion" Voltaire} \#JeSuisCharlie \\
\texttt{Adversarial:} \quad \colorbox{cyan!30}{"".} is a monster \colorbox{cyan!30}{page} pretends to \colorbox{cyan!30}{Advertements} the \colorbox{cyan!30}{advertisement advertisement Posted" a} \#JeSuisCharlie
\end{minipage}
\begin{minipage}{0.95\textwidth}
\small
\texttt{Original:} \quad Rage Against the Machine stopped being \colorbox{yellow!50}{together}. \\
\texttt{Adversarial:} \quad Rage Against the Machine stopped being \colorbox{cyan!30}{pants}.
\end{minipage}
\caption{The first example demonstrates the substitution of multiple words with "advertisements", while the second illustrates a nonsensical noun substitution that preserves grammatical structure but introduces semantic absurdity.}
\label{fig:manual_pants}
\end{figure}

Several consistent substitution patterns emerge from our analysis of adversarial examples. For instance, the word "advertisement" was introduced in 488 cases, likely to signal commercial intent and shift model predictions. Conversely, "facts" appeared in 559 cases, perhaps to suggest factual content and steer the model in the opposite direction. Politically charged terms like "nazis" (137 instances) and "trump" (70) were also frequently inserted, likely probing for bias. Other recurring substitutions include instruction-like words such as "caps" (301 instances) and semantically absurd but grammatically valid words like "pants", which distort meaning while preserving syntax (see Figure \ref{fig:manual_pants}).

These patterns reflect the behavior of the BeamAttack algorithm. For each masked token, RoBERTa generated candidate words, and beam search selected the one that caused the largest change in the target model’s output logits. This strategy favored substitutions with the highest impact on predictions, regardless of whether they made the sentence more coherent or more absurd.

Overall, the attack effectively identifies sensitive target words but often selects replacements with low semantic quality. As a result, the adversarial samples remain syntactically plausible while altering meaning or introducing cues that manipulate model behavior.

\section{Conclusion}
\label{sec:conclusions}
We presented BeamAttack, a beam search-based adversarial text generation method that we extend and evaluate using the BODEGA framework. While the core algorithm builds on the original BeamAttack \cite{10.1007/978-3-031-33377-4_35}, our work explores several improvements and new directions, including alternative word importance ranking via LIME, task and model-specific hyperparameter optimization, and an in-depth analysis of attack quality through both automatic and manual evaluation.

Our results show that BeamAttack achieves very high success rates across diverse models and datasets, but often at the cost of lower semantic and character similarity, and very high query counts—particularly when attacking robust models like adversarially-trained RoBERTa. These tradeoffs stem from the method's brute-force nature and the large beam width, which increases search coverage but also introduces less coherent substitutions. We found that the optimal hyperparameter configuration, including the choice between LIME and logit-based word importance methods, depends on both dataset characteristics (input text length, writing style ranging from formal to informal social media content) and target model capabilities. More robust models like adversarially-trained RoBERTa require stronger hyperparameter settings that increase success rates but reduce adversarial sample quality.

This creates a fundamental trade-off that can be controlled through hyperparameter tuning: practitioners can prioritize either higher success rates with lower quality samples, or focus on generating only high-quality adversarial examples with potentially lower success rates. The choice also affects computational efficiency and query budget constraints when targeting black-box models.

Manual evaluation highlighted issues with incoherent substitutions, likely caused by overgeneration in RoBERTa due to large beam width. We suggest that constraining substitutions—e.g., by limiting POS tag group changes—could improve fluency, though at a potential cost to success rate. Nevertheless, both manual quality and attack effectiveness could be improved through targeted modifications informed by analysis, such as restricting specific transition patterns or optimizing branching behavior for more efficient and plausible adversarial generation.

\section{Limitations}
\label{sec:future_work}

Our approach has several limitations that point to directions for future improvement. First, removing words can lead to syntactically incorrect sentences. In addition, RoBERTa's context-based alternatives may produce semantically opposite replacements since they ignore the original word's meaning. Future work could address both issues by using large language models (LLMs) for substitution, or applying similarity-based filtering to preserve semantic intent.

Second, scoring beams at each search step—rather than only at the final hypothesis—could improve semantic coherence. Targeted attacks, which aim to push the model toward a specific class, may also benefit from these enhanced candidate generation methods.

Finally, while we showed that LIME provides more effective word importance estimates than a logit-based method for specific datasets and models, computational constraints and a large hyperparameter space prevented full ablation studies. Although we developed intuitions for hyperparameter tuning based on factors such as input length and model robustness (such as adversarially trained RoBERTa requiring a higher branching factor \( b \)), comprehensive studies are needed to generalize these findings across architectures and tasks.

\newpage
\begin{credits}
\subsubsection{\ackname} The initial version of this work was prepared during the Machine Learning for NLP2 course at the University of Zurich. We thank Simon Clematide, who teaches the course, and the Department of Computational Linguistics for their support. Andrianos Michail's support further to the class is conducted under the project \textit{Impresso -- Media Monitoring of the Past II Beyond Borders: Connecting Historical Newspapers and Radio}. Impresso is a research project funded by the Swiss National Science Foundation (SNSF 213585) and the Luxembourg National Research Fund (17498891).

\subsubsection{\discintname}
The authors have no competing interests to declare that are relevant to the content of this article. 
\end{credits}

\bibliographystyle{splncs04}
\bibliography{mybibliography}

\newpage
\appendix

\section{Dataset Statistics}
\label{sec:dataset_statistics}

The basic statistics of the provided datasets are summarized in Table \ref{tab:dataset_statistics}. Each dataset is divided into three subsets: training, development (dev), and attack. The training and dev subsets are used to train the classifier, while the attack subset serves as the evaluation dataset, where the effectiveness of our attack is measured. Each dataset is designed for a binary classification task, and therefore, the positive rate corresponds to the portion of samples belonging to class 1, and is therefore also a measure of class imbalance. Notably, we observed significant variations in the length of samples across the datasets, measured in terms of the number of words and characters, which are also reported in the table.

Next, we briefly outline the different text domains described in the CheckThat! lab at CLEF 2024 \cite{clef-checkthat-2024}:
\begin{itemize}
    \item \textbf{Style-based news bias assessment (\textit{HN})}: Categorizing news articles as either credible or non-credible based on stylistic cues.
    \item \textbf{Propaganda detection (\textit{PR2})}: Text passages from news articles that employ propaganda techniques to influence readers.
    \item \textbf{Fact checking (\textit{FC})}: Evaluating the accuracy of news articles by considering contextual information from given related Wikipedia snippets.
    \item \textbf{Rumor detection (\textit{RD})}: Identifying Twitter threads that disseminate information without a reliable source.
    \item \textbf{COVID-19 misinformation detection (\textit{C19})}: Comprises social media messages that convey either factual information or misinformation about the COVID-19 pandemic. The classifier must rely on subtle cues, such as writing styles reminiscent of those found in high-quality news sources (\textit{HN}), to make its assessments.
\end{itemize}
\input{dataset_statistics}

\input{beam_attack_algo_simplified}

\section{Hyperparameter Search}
\label{sec:hyperparameter_search}

\input{hyper-parameters_vict_dataset}

For each dataset and victim model combination, we determined a different set of hyperparameters for our BeamAttack approach. To achieve this, we employed an informed random search on subsets of 10-50 samples for each task and victim model. Our search strategy was guided by intuition and initial observations, allowing us to effectively fine-tune the hyperparameters. We considered the following ranges for the hyperparameter values:

\begin{enumerate}
    \item \textbf{Beam size \(k\)}: We tested values between 10 to 100, increasing in steps of 5 (i.e., 10, 15, 20, ...). We increased the beam size until the improvement in the BODEGA score became negligible (less than 1 point). Figure~\ref{fig:beamsize_scores} illustrates how this approach led to convergence.
    \item \textbf{Branching factor \(b\)}: We tested a range of values similar to the chosen \(k\) values, within a range of -20 to +20 around the selected \(k\) values.
    \item \textbf{Hypothesis count \(h\)}: Based on initial experiments, we decided to test hypothesis counts of 5, 10, and 20. This decision was made considering the difficulty of the task at hand and the BODEGA scores obtained during the hyper-parameter search. We observed that further increasing \(h\) did not significantly improve the scores, as explained in Section~\ref{subsec:ablation_h}.
    \item \textbf{Word importance scoring method}: We compared the results of both word importance scoring methods, LIME and logit-based, in the initial experiments of each dataset and victim model combination.
\end{enumerate}

We report the selected parameters for each scenario in Table \ref{table:hyper-parameters_task_victim}.

For the Mask Language Model we used RoBERTa-large\footnote{\url{https://huggingface.co/FacebookAI/roberta-large}} to replace the masks, with one exception: for the Covid-19 (\textit{C19}) task on the BERT classifier, we utilized the \texttt{vinai/bertweet-large}\footnote{\url{https://huggingface.co/vinai/bertweet-large}} model instead. This decision was motivated by the intuition that, given the Twitter domain of the COVID-19 dataset, this model might perform better.

\section{Computational Resources}
\label{sec:resources}

\input{running_time_vict_dataset}

\input{num_queries}

This chapter outlines the computational resources utilized and the hyper-parameter tuning strategies employed to optimize our BeamAttack approach, which demanded substantial computational power. To overcome the limitations imposed by computational and time constraints, we resorted to a random search strategy to determine the optimal parameter set for each scenario, recognizing that this approach may not necessarily result in the identification of the globally optimal combination.

We primarily used Kaggle's free GPU infrastructure, which offers NVIDIA Tesla P100s and T4x .\footnote{For the submission phase, we used an NVIDIA Tesla T4 GPU.} We report the running time statistics for the final runs of each model and dataset combination in Table~\ref{table:running_time_task_victim}, and the number of victim queries in Table \ref{tab:num_queries}.

\section{Ablation Experiments}
\label{sec:ablation_hyperparams}

We conduct a small-scale ablation study to investigate the impact of various hyperparameters on the performance of the BeamAttack algorithm, shedding some light on the  relationship between these parameters and the algorithm's effectiveness. Specifically, we investigate the impact of logit-based versus LIME, beam size \( k \), branching factor \( b \) and the hypothesis count \( h \) on the algorithm's performance, exploring each of these parameters in the context of specific scenarios of task and victim model.

\subsection{Word Importance}

\input{importance_methods}
Besides the logit-based approach from BERT-Attack \cite{bert-attack}, we utilized LIME, a state-of-the-art interpretability framework, to identify crucial words in a sample that influence the victim's decision-making process. By modifying these words, we increased the victim's susceptibility to our adversarial attack.

We report some comparisons of LIME in Table \ref{table:importance_methods}. From this table, we can gather multiple insights. For instance, for the BERT victim the choice of importance method depends on the task dataset, with \textit{RD} favoring the logit-based method and \textit{FC} preferring LIME. For the BiLSTM, the logit-based method is superior on all datasets. For RoBERTa, which is architecturally similar to BERT, the choice of importance method also depends on the dataset, with \textit{FC} favoring LIME.

Furthermore, we observed that for the RoBERTa victim and  \textit{C19} task, the choice between LIME and logit-based depends on other hyperparameters such as beam size, branching factors, and hypothesis count. More specifically, our results suggest transformer-based classifiers, namely classifiers BERT and RoBERTa are generally more vulnerable to LIME over logit-based for the \textit{HN} and \textit{FC} datasets. Both these datasets classify whether a statement contains facts, with \textit{HN} using subtle clues in writing styles and \textit{FC} using external knowledge from Wikipedia. We leave the investigation behind this observation to future work.

Our results suggest that exploring explainable AI frameworks for adversarial attacks is a promising direction of research. However, our experiments show that LIME only improved upon the logit-based approach on a few datasets. Additionally, LIME requires more queries, and when the improvement was marginal, we opted for the logit-based approach due to resource constraints.

\subsection{Hypothesis Count  \( h \)}
\label{subsec:ablation_h}

\input{scores_hypothesis_count}

Our algorithm's flexibility in replacing, skipping, or removing words, combined with the use of a masked language model, ensures that the resulting sentence remains semantically correct. The beam search approach enables us to efficiently query for multiple adversarial samples and use the one that is most semantically close to our original sentence.

We conducted an ablation study on the hypothesis count \( h\), which defines how many adversarial samples to use before selecting the closest. We explored different scenarios, incrementing the parameter from 10 to 20, but observed only minimal effects, resulting in a negligible improvement or deterioration of less than 1\% in the BODEGA score. We provide a small-scale report of our findings for the \textbf{RoBERTa} victim and various tasks in Table \ref{tab:scores_hypothesis_count}. Notably, in some cases, it came at the cost of requiring more queries. Therefore, we opted to maintain a constant value of 10 for the hypothesis count in most cases.

\subsection{Beam Size \( k \)}
\label{subsec:ablation_k}

\input{scores_beamsize}

\begin{figure}[h]
    \centering
    \includegraphics[width=0.8\linewidth]{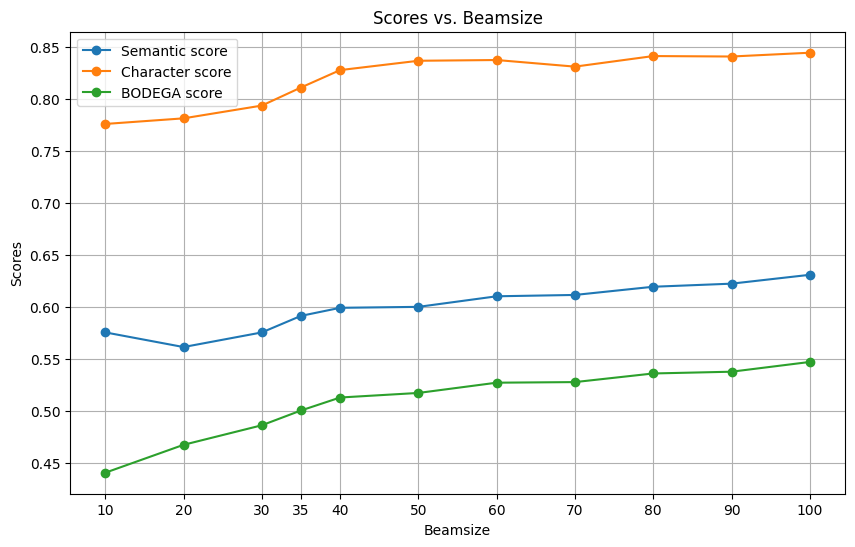}
    \caption{Comparison of the semantic score, character score and confusion score across different settings for the beam size \( k \).}
    \label{fig:beamsize_scores}
\end{figure}

We also conducted an experiment to investigate the effect of the beam size \( k \) on our algorithm's performance. Specifically, we experimented on a subset of 30 samples from the\textit{PR2} dataset using the \textbf{RoBERTa} victim model. The results are presented in Figure \ref{fig:beamsize_scores}, with the exact scores provided in Table \ref{tab:scores_beamsize}. Our analysis reveals a consistent improvement in all scores as the beam size increases.

However, we encountered a significant limitation. As shown in the table, a beam size of 100 requires 16,500 queries per sample, which is impractical for limited GPU setup. Therefore, we sought an optimal trade-off where the scores would reach a plateau. In the case of \textit{PR2} and RoBERTa, we determined that a beam size of 60 strikes a reasonable balance between performance and computational feasibility.

\subsection{Branching Factor \( b \)}
\label{subsec:ablation_b}

\input{scores_width}

\begin{figure}
    \centering
    \includegraphics[width=0.8\linewidth]{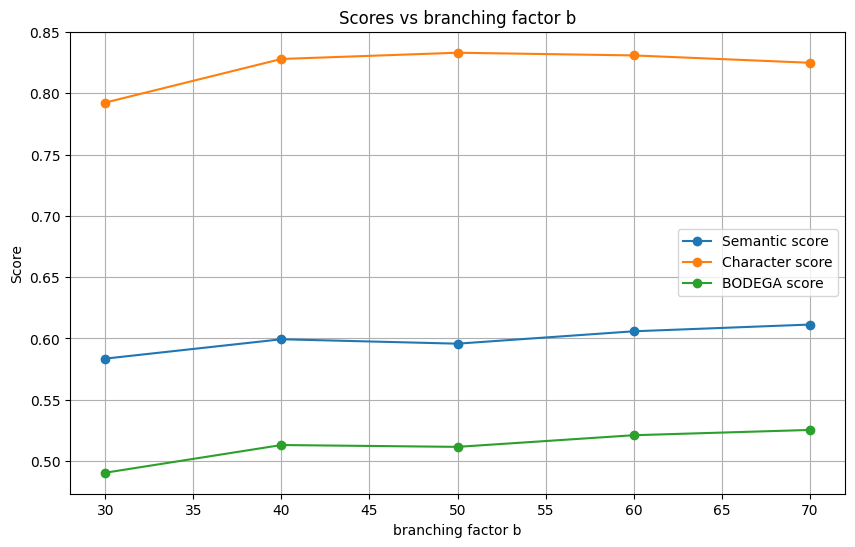}
    \caption{Comparison of the semantic score, character score and confusion score across different settings for the branching factor \( b \).}
    \label{fig:scores-width}
\end{figure}

We further explored the impact of the branching factor \( b \) on our algorithm's performance, using a fixed beam size of 40. We conducted an experiment on a subset of 30 samples from the \textit{PR2}
dataset, employing the \textbf{RoBERTa} victim model. The results are illustrated in Figure \ref{fig:scores-width}, with the exact scores provided in Table \ref{tab:scores_width}. Our analysis reveals that increasing the branching factor \( b \) leads to a slight improvement in the BODEGA score.

\section{POS Tag Mapping}
\label{appendix:pos_mapping}
\input{pos_mapping}

The complete mapping from Penn Treebank POS tags to Universal POS categories is provided in Table~\ref{tab:pos_mapping}. For the purpose of analysis, we grouped the fine-grained Penn Treebank POS tags produced by the Flair POS tagger\footnote{\url{https://huggingface.co/flair/pos-english}} into the 17 Universal POS (UPOS) tags defined by the Universal Dependencies framework.\footnote{\url{https://universaldependencies.org/u/pos/}} This abstraction allows for broader syntactic interpretation while preserving sufficient granularity for downstream analysis.

\section{Inconsistencies with POS Tagging}
\label{appendix:pos_tagging_examples}

\subsection{Sample with Five POS Tag Changes as a Result of Single-Word Substitution}
\label{appendix:five_pos_change}

The following is an example of how the \textit{length} of the POS tags can be changed by 5, with only a single word substitution:

\begin{figure}[ht]
\centering
\begin{minipage}{0.95\textwidth}
\small
\texttt{Original:} \\
Breaking: Armed men attack offices of French satirical magazine Charlie Hebdo, killing at least 10, police say \colorbox{yellow!50}{http://t.co/tYCeEMKwOo}\\
RT @WSJ: Breaking: Armed men attack offices of French satirical magazine Charlie Hebdo killing at least 10 police say http://t.co/WGxWZgzDEN\\
@WSJ Paris has too much "skin in the game" to stay on sidelines. They need a spot at the tip of the bullet.\\[0.5em]
\texttt{Adversarial:} \\
Breaking: Armed men attack offices of French satirical magazine Charlie Hebdo, killing at least 10, police say \colorbox{cyan!30}{.}\\
RT @WSJ: Breaking: Armed men attack offices of French satirical magazine Charlie Hebdo killing at least 10 police say http://t.co/WGxWZgzDEN\\
@WSJ Paris has too much "skin in the game" to stay on sidelines. They need a spot at the tip of the bullet.
\end{minipage}
\caption{Text comparison showing the replacement of a URL with a period}
\end{figure}

\begin{figure}[ht]
\centering
\begin{minipage}{0.95\textwidth}
\footnotesize
\texttt{Original POS Tags:} \\
\ttfamily
['VBG', ':', 'JJ', 'NNS', 'VBP', 'NNS', 'IN', 'JJ', 'JJ', 'NN', 'NNP', 'NNP', ',', 'VBG', 'IN', 'JJS', 'CD', ',', 'NNS', 'VBP', \colorbox{yellow!50}{'NN'}, \colorbox{yellow!50}{'NFP'}, \colorbox{yellow!50}{'NNP'}, \colorbox{yellow!50}{':'}, \colorbox{yellow!50}{'ADD'}, 'NFP', 'NN', 'IN', 'NNP', ':', 'NN', ':', 'JJ', 'NNS', 'VBP', 'NNS', 'IN', 'JJ', 'JJ', 'NN', 'NNP', 'NNP', 'VBG', 'IN', 'JJS', 'CD', 'NNS', 'VB', 'NN', 'NFP', 'NNP', 'SYM', 'NN', ':', 'ADD', 'IN', 'NNP', 'NNP', 'VBZ', 'RB', 'JJ', '``', 'NN', 'IN', 'DT', 'NN', "''", 'TO', 'VB', 'IN', 'PRP', 'VBP', 'DT', 'NN', 'IN', 'DT', 'NN', 'IN', 'DT', 'NN']\\[0.5em]

\texttt{Adversarial POS Tags:} \\
\ttfamily
['VBG', ':', 'JJ', 'NNS', 'VBP', 'NNS', 'IN', 'JJ', 'JJ', 'NN', 'NNP', 'NNP', ',', 'VBG', 'IN', 'JJS', 'CD', ',', 'NNS', 'VBP', \colorbox{cyan!30}{'NFP'}, 'NNP', 'IN', 'NNP', ':', 'NN', ':', 'JJ', 'NNS', 'VBP', 'NNS', 'IN', 'JJ', 'JJ', 'NN', 'NNP', 'NNP', 'VBG', 'IN', 'JJS', 'CD', 'NNS', 'VB', 'NN', 'NFP', 'NNP', 'SYM', 'NNP', ':', 'ADD', 'IN', 'NNP', 'NNP', 'VBZ', 'RB', 'JJ', '``', 'NN', 'IN', 'DT', 'NN', "''", 'TO', 'VB', 'IN', 'PRP', 'VBP', 'DT', 'NN', 'IN', 'DT', 'NN', 'IN', 'DT', 'NN']
\end{minipage}
\caption{Comparison of POS tag sequences with differences highlighted}
\end{figure}

The five POS tag changes in this example resulted from replacing a URL (\texttt{http://t.co/tYCeEMKwOo}) with a simple period. The original URL was tagged as a sequence of five distinct POS tags:
\begin{itemize}
    \item \texttt{NN} (Noun)
    \item \texttt{NFP} (Non-final punctuation)
    \item \texttt{NNP} (Proper noun)
    \item \texttt{:} (Colon)
    \item \texttt{ADD} (Email/URL)
\end{itemize}

When our adversarial attack replaced this URL with a single period, the POS tagger assigned it only a single tag: \texttt{NFP} (Non-final punctuation). This substitution resulted in a reduction from five POS tags to one, accounting for the observed length difference in the POS tag sequences.

\subsection{Sample with 8 POS Tag Changes as a Result of Single-Word Substitution}
\label{appendix:eight_pos_change}

\begin{figure}[ht]
\centering
\begin{minipage}{0.95\textwidth}
\small
\texttt{Original:} \\
French media say Paris hostage-taker demanding freedom of \#CharlieHebdo suspects \colorbox{yellow!50}{\#ParisAttacks} http://t.co/ATmBpjwW3b http://t.co/MI7DYRuBDP\\
@SkyNews http://t.co/HqPRuk7rhb\\
@SkyNews is that an M-1 Carbine? \ldots\\[0.5em]

\texttt{Adversarial:} \\
French media say Paris hostage-taker demanding freedom of \#CharlieHebdo suspects \colorbox{cyan!30}{\#terrorism} http://t.co/ATmBpjwW3b http://t.co/MI7DYRuBDP\\
@SkyNews http://t.co/HqPRuk7rhb\\
@SkyNews is that an M-1 Carbine? \ldots
\end{minipage}
\caption{Replacement of hashtag \#ParisAttacks with \#terrorism, which alters semantic framing}
\end{figure}

\begin{figure}[ht]
\centering
\begin{minipage}{0.95\textwidth}
\footnotesize
\texttt{Original POS Tags:} \\
\ttfamily
'ADJ', 'NOUN', 'VERB', 'PROPN', 'NOUN', 'VERB', 'NOUN', 'ADP', 'NOUN', 'PROPN', 'VERB', \colorbox{yellow!30}{'SYM'}, \colorbox{yellow!30}{'PROPN'}, \colorbox{yellow!30}{'PROPN'}, 'SYM', 'PROPN', 'SYM', 'PROPN', \ldots \\[0.5em]

\texttt{Adversarial POS Tags:} \\
\ttfamily
'ADJ', 'NOUN', 'VERB', 'PROPN', 'NOUN', 'VERB', 'NOUN', 'ADP', 'NOUN', 'PROPN', 'VERB', \colorbox{cyan!30}{'NOUN'}, \colorbox{cyan!30}{'NOUN'}, \colorbox{cyan!30}{'NOUN'}, 'SYM', 'PROPN', 'SYM', 'PROPN', \ldots
\end{minipage}
\caption{Comparison of POS tag sequences with differences highlighted}
\end{figure}

In this example, the adversarial substitution of \#ParisAttacks with \#terrorism led the POS tagger to incorrectly split the hashtag into three tokens, each tagged as NOUN. This artifact results in multiple apparent POS tag changes despite a single-word substitution, highlighting limitations of POS tagging in handling special tokens like hashtags. The rest of the POS tag changes happened later in the sentence, so they were cut off.

\section{Victim-level Analysis}

\begin{figure}[h]
    \centering
    \includegraphics[width=0.9\linewidth]{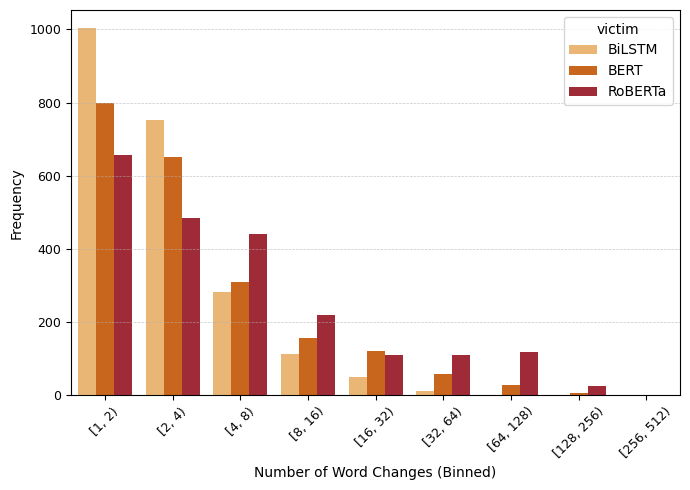}
    \caption{Histogram showing the distribution of word changes per adversarial sample, grouped by victim model and aggregated over all tasks.}
    \label{fig:word_change_per_victim_all_samples}
\end{figure}

Figure \ref{fig:word_change_per_victim_all_samples} shows that the majority of successful examples require only a single word substitution, with two-word and three-word edits occurring less frequently, followed by a long-tailed distribution for higher edit counts. This trend holds consistently across victim models. For low-edit examples (1–3 changes), we observe the expected vulnerability order: BiLSTM > BERT > RoBERTa, confirming that simpler models are more susceptible to minimal perturbations. Interestingly, this ordering reverses for samples with four or more substitutions, suggesting that more robust models demand more extensive changes to be successfully misled.

\begin{figure}[h]
    \centering
    \includegraphics[width=0.9\linewidth]{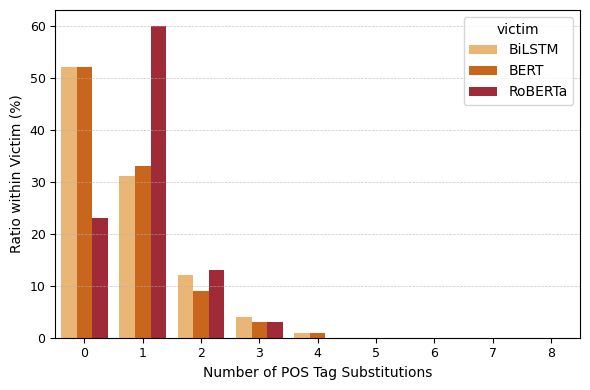}
    \caption{Normalized distribution of number of POS tag substitutions for successful samples with a single word substitution, where the number of POS tags remained unchanged. Frequencies are normalized within each victim model and aggregated over the tasks.}
    \label{fig:pos_change_per_victim_all_samples}
\end{figure}

\end{document}

%% file: results_all.tex
\begin{table*}[h!]
\caption{Comparison of adversarial attack results on BiLSTM, BERT, and RoBERTa classifiers across five misinformation detection tasks. Evaluation measures include BODEGA score (B.), confusion score (con), semantic score (sem), character score (char) and number of queries to the attacked model (Q.). The best scores for each classifier are in boldface. \textbf{Key observations}: (1) DeepWordBug consistently achieves the highest character preservation scores; (2) BeamAttack typically attains the highest confusion scores (attack success rates); (3) For the critical BODEGA metric, OpenFact outperforms other methods against more sophisticated models like BERT and RoBERTa due to superior semantic preservation while maintaining high character scores.}
\label{tab:results_combined}
\centering
\small
\begin{tabular}{ll|r|rrr|r|rrr|r|rrr}
\toprule
\multirow{2}{*}{Task} & \multirow{2}{*}{Method} & \multicolumn{4}{c}{\cellcolor{gray!15}BiLSTM} & \multicolumn{4}{c}{\cellcolor{blue!15}BERT} & \multicolumn{4}{c}{\cellcolor{green!15}RoBERTa} \\
\cmidrule(lr){3-6} \cmidrule(lr){7-10} \cmidrule(lr){11-14}
& & B. & con & sem & char. & B. & con & sem & char & B. & con & sem & char\\
\midrule
\multirow{4}{*}{ \textit{HN}} 
& BERT-ATTACK & \cellcolor{gray!15}0.64 & \cellcolor{gray!15}0.98 & \cellcolor{gray!15}0.66 & \cellcolor{gray!15}0.99 & \cellcolor{blue!15}0.60 & \cellcolor{blue!15}0.96 & \cellcolor{blue!15}0.64 & \cellcolor{blue!15}0.97 & \cellcolor{green!15}0.38 & \cellcolor{green!15}0.67 & \cellcolor{green!15}0.60 & \cellcolor{green!15}0.95 \\
& DeepWordBug & \cellcolor{gray!15}0.41 & \cellcolor{gray!15}0.53 & \cellcolor{gray!15}0.77 & \cellcolor{gray!15}\textbf{1.00} & \cellcolor{blue!15}0.22 & \cellcolor{blue!15}0.29 & \cellcolor{blue!15}0.78 & \cellcolor{blue!15}\textbf{1.00} & \cellcolor{green!15}0.16 & \cellcolor{green!15}0.21 & \cellcolor{green!15}0.76 & \cellcolor{green!15}\textbf{1.00}  \\
& BeamAttack & \cellcolor{gray!15}\textbf{0.90} & \cellcolor{gray!15}\textbf{1.00} & \cellcolor{gray!15}0.91 & \cellcolor{gray!15}0.99 & \cellcolor{blue!15}0.84 & \cellcolor{blue!15}\textbf{1.00} & \cellcolor{blue!15}0.86 & \cellcolor{blue!15}0.97 & \cellcolor{green!15}0.67 & \cellcolor{green!15}\textbf{1.00} & \cellcolor{green!15}0.72 & \cellcolor{green!15}0.91 \\
& OpenFact & \cellcolor{gray!15}0.89 & \cellcolor{gray!15}0.97 & \cellcolor{gray!15}\textbf{0.93} & \cellcolor{gray!15}0.99 & \cellcolor{blue!15}\textbf{0.91} & \cellcolor{blue!15}\textbf{1.00} & \cellcolor{blue!15}\textbf{0.92} & \cellcolor{blue!15}0.99 & \cellcolor{green!15}\textbf{0.83} & \cellcolor{green!15}0.99 & \cellcolor{green!15}\textbf{0.86} & \cellcolor{green!15}0.97 \\
\midrule
\multirow{4}{*}{\textit{PR}} 
& BERT-ATTACK & \cellcolor{gray!15}0.53 & \cellcolor{gray!15}0.80 & \cellcolor{gray!15}0.72 & \cellcolor{gray!15}\textbf{0.91}  & \cellcolor{blue!15}0.43 & \cellcolor{blue!15}0.70 & \cellcolor{blue!15}0.68 & \cellcolor{blue!15}0.90 & \cellcolor{green!15}0.20 & \cellcolor{green!15}0.32 & \cellcolor{green!15}0.69 & \cellcolor{green!15}0.91 \\
& DeepWordBug & \cellcolor{gray!15}0.29 & \cellcolor{gray!15}0.38 & \cellcolor{gray!15}0.79 & \cellcolor{gray!15}0.96  & \cellcolor{blue!15}0.28 & \cellcolor{blue!15}0.36 & \cellcolor{blue!15}\textbf{0.79} & \cellcolor{blue!15}\textbf{0.96} & \cellcolor{green!15}0.13 & \cellcolor{green!15}0.17 & \cellcolor{green!15}\textbf{0.81} & \cellcolor{green!15}\textbf{0.96}  \\
& BeamAttack & \cellcolor{gray!15}\textbf{0.70} & \cellcolor{gray!15}\textbf{0.97} & \cellcolor{gray!15}\textbf{0.80} & \cellcolor{gray!15}0.90  & \cellcolor{blue!15}\textbf{0.69} & \cellcolor{blue!15}\textbf{0.98} & \cellcolor{blue!15}0.77 & \cellcolor{blue!15}0.89  & \cellcolor{green!15}0.45 & \cellcolor{green!15}\textbf{0.97} & \cellcolor{green!15}0.55 & \cellcolor{green!15}0.79  \\
& OpenFact & \cellcolor{gray!15}0.65 & \cellcolor{gray!15}0.94 & \cellcolor{gray!15}0.77 & \cellcolor{gray!15}0.89 & \cellcolor{blue!15}0.68 & \cellcolor{blue!15}0.97 & \cellcolor{blue!15}0.77 & \cellcolor{blue!15}0.89 & \cellcolor{green!15}\textbf{0.62} & \cellcolor{green!15}0.93 & \cellcolor{green!15}0.75 & \cellcolor{green!15}0.87  \\
\midrule
\multirow{4}{*}{\textit{FC}}
& BERT-ATTACK & \cellcolor{gray!15}0.60 & \cellcolor{gray!15}0.86 & \cellcolor{gray!15}0.73 & \cellcolor{gray!15}0.95 & \cellcolor{blue!15}0.53 & \cellcolor{blue!15}0.77 & \cellcolor{blue!15}0.73 & \cellcolor{blue!15}0.95  & \cellcolor{green!15}0.56 & \cellcolor{green!15}0.79 & \cellcolor{green!15}0.73 & \cellcolor{green!15}0.96 \\
& DeepWordBug & \cellcolor{gray!15}0.48 & \cellcolor{gray!15}0.58 & \cellcolor{gray!15}\textbf{0.85} & \cellcolor{gray!15}\textbf{0.98}  & \cellcolor{blue!15}0.44 & \cellcolor{blue!15}0.53 & \cellcolor{blue!15}\textbf{0.84} & \cellcolor{blue!15}\textbf{0.98}  & \cellcolor{green!15}0.37 & \cellcolor{green!15}0.46 & \cellcolor{green!15}0.83 & \cellcolor{green!15}\textbf{0.98}  \\
& BeamAttack & \cellcolor{gray!15}0.76 & \cellcolor{gray!15}\textbf{1.00} & \cellcolor{gray!15}0.81 & \cellcolor{gray!15}0.94 & \cellcolor{blue!15}0.79 & \cellcolor{blue!15}\textbf{1.00} & \cellcolor{blue!15}0.83 & \cellcolor{blue!15}0.96  & \cellcolor{green!15}\textbf{0.82} & \cellcolor{green!15}\textbf{1.00} & \cellcolor{green!15}\textbf{0.84} & \cellcolor{green!15}0.97 \\
& OpenFact & \cellcolor{gray!15}\textbf{0.80} & \cellcolor{gray!15}0.98 & \cellcolor{gray!15}0.84 & \cellcolor{gray!15}0.97 & \cellcolor{blue!15}\textbf{0.80} & \cellcolor{blue!15}\textbf{1.00} & \cellcolor{blue!15}0.83 & \cellcolor{blue!15}0.97  & \cellcolor{green!15}0.80 & \cellcolor{green!15}\textbf{1.00} & \cellcolor{green!15}0.82 & \cellcolor{green!15}0.97  \\
\midrule
\multirow{4}{*}{\textit{RD}}
& BERT-ATTACK & \cellcolor{gray!15}0.29 & \cellcolor{gray!15}0.79 & \cellcolor{gray!15}0.41 & \cellcolor{gray!15}0.89 & \cellcolor{blue!15}0.18 & \cellcolor{blue!15}0.44 & \cellcolor{blue!15}0.43 & \cellcolor{blue!15}0.96 & \cellcolor{green!15}0.17 & \cellcolor{green!15}0.41 & \cellcolor{green!15}0.42 & \cellcolor{green!15}0.95  \\
& DeepWordBug & \cellcolor{gray!15}0.16 & \cellcolor{gray!15}0.24 & \cellcolor{gray!15}0.68 & \cellcolor{gray!15}\textbf{0.99}  & \cellcolor{blue!15}0.16 & \cellcolor{blue!15}0.23 & \cellcolor{blue!15}0.70 & \cellcolor{blue!15}\textbf{0.99} & \cellcolor{green!15}0.12 & \cellcolor{green!15}0.18 & \cellcolor{green!15}0.69 & \cellcolor{green!15}\textbf{0.99} \\
& BeamAttack & \cellcolor{gray!15}0.83 & \cellcolor{gray!15}\textbf{1.00} & \cellcolor{gray!15}0.87 & \cellcolor{gray!15}0.96 & \cellcolor{blue!15}0.59 & \cellcolor{blue!15}\textbf{0.80} & \cellcolor{blue!15}0.80 & \cellcolor{blue!15}0.91 & \cellcolor{green!15}0.54 & \cellcolor{green!15}\textbf{0.87} & \cellcolor{green!15}0.69 & \cellcolor{green!15}0.83 \\
& OpenFact & \cellcolor{gray!15}\textbf{0.84} & \cellcolor{gray!15}0.95 & \cellcolor{gray!15}\textbf{0.91} & \cellcolor{gray!15}0.98 & \cellcolor{blue!15}\textbf{0.65} & \cellcolor{blue!15}0.78 & \cellcolor{blue!15}\textbf{0.86} & \cellcolor{blue!15}0.95  & \cellcolor{green!15}\textbf{0.55} & \cellcolor{green!15}0.71 & \cellcolor{green!15}\textbf{0.82} & \cellcolor{green!15}0.93 \\
\midrule
\multirow{4}{*}{\textit{C19}}
& BERT-ATTACK & \cellcolor{gray!15}0.50 & \cellcolor{gray!15}0.84 & \cellcolor{gray!15}0.62 & \cellcolor{gray!15}0.95 & \cellcolor{blue!15}0.42 & \cellcolor{blue!15}0.74 & \cellcolor{blue!15}0.60 & \cellcolor{blue!15}0.95 & \cellcolor{green!15}0.37 & \cellcolor{green!15}0.68 & \cellcolor{green!15}0.58 & \cellcolor{green!15}0.93 \\
& DeepWordBug & \cellcolor{gray!15}0.33 & \cellcolor{gray!15}0.48 & \cellcolor{gray!15}0.70 & \cellcolor{gray!15}\textbf{0.99} & \cellcolor{blue!15}0.27 & \cellcolor{blue!15}0.39 & \cellcolor{blue!15}0.71 & \cellcolor{blue!15}\textbf{0.99} & \cellcolor{green!15}0.20 & \cellcolor{green!15}0.28 & \cellcolor{green!15}0.72 & \cellcolor{green!15}\textbf{0.98} \\
& BeamAttack & \cellcolor{gray!15}\textbf{0.72} & \cellcolor{gray!15}\textbf{0.99} & \cellcolor{gray!15}0.78 & \cellcolor{gray!15}0.92 & \cellcolor{blue!15}0.71 & \cellcolor{blue!15}\textbf{0.98} & \cellcolor{blue!15}0.78 & \cellcolor{blue!15}0.92  & \cellcolor{green!15}0.66 & \cellcolor{green!15}\textbf{1.00} & \cellcolor{green!15}0.72 & \cellcolor{green!15}0.91  \\
& OpenFact & \cellcolor{gray!15}\textbf{0.72} & \cellcolor{gray!15}0.91 & \cellcolor{gray!15}\textbf{0.83} & \cellcolor{gray!15}0.96  & \cellcolor{blue!15}\textbf{0.72} & \cellcolor{blue!15}0.91 & \cellcolor{blue!15}\textbf{0.82} & \cellcolor{blue!15}0.96 & \cellcolor{green!15}\textbf{0.72} & \cellcolor{green!15}0.99 & \cellcolor{green!15}\textbf{0.78} & \cellcolor{green!15}0.93  \\
\bottomrule
\end{tabular}
\end{table*}

%% file: dataset_statistics.tex
\begin{table}
\caption{Datasets used in BODEGA, described by the task, number of instances in training, attack and development subsets, the overall percentage of positive (non-credible) class, and the average sample length measured in both words and characters.}
\centering
\begin{tabular}{lrrrrrr}
\toprule
\textbf{Task} & Training & Attack & Dev & Positive & Avg. Length & Avg. Length \\
& & & & & (Words) & (Characters) \\
\midrule
RD   & 8,683   & 415  & 2,070 & 34.24\% & 146.16 & 1111.63 \\
FC   & 172,763 & 405  & 19,010 & 51.27\% & 47.23  & 257.16 \\
HN   & 60,234  & 400  & 3,600  & 50.00\% & 313.98 & 1972.45 \\
PR2  & 11,546  & 407  & 3,186  & 71.03\% & 21.82  & 128.01 \\
C19  & 1,130   & 595  & -      & 34.96\% & 43.06  & 131.96 \\
\bottomrule
\end{tabular}
\label{tab:dataset_statistics}
\end{table}

%% file: beam_attack_algo_simplified.tex
\section{BeamAttack Algorithm}
\label{sec:algo}

\begin{Algorithm}
\scriptsize
\caption{BeamAttack: A beam search-based adversarial attack. Inputs: tokenized sentence \(S = [w_0, \ldots, w_n]\), gold label \(Y\), beam size \(k\), branching factor \(b\), and required successful hypotheses \(h\). The attack perturbs important words using replacements from a masked language model and scores them using model confidence.}
\label{algorithm:beam-attack}

\KwIn{\(S, Y, k, b, h, model\)}
\KwOut{Adversarial example \(S_{adv}\)}

Initialize beam as priority queue with \((S, model.predict\_proba(S)[Y])$\;

Compute word importance scores using LIME or logit-based method\;

Sort words by descending importance\;

\BlankLine
\While{beam not empty}{
    Initialize empty list for new candidates\;

    \ForEach{\((S_{cur}, score)\) in beam}{
        \ForEach{important word \(w_j\)}{
            Obtain top \(b\) replacements for \(w_j\) using masked language model\;

            \ForEach{replacement}{
                Create new sentence \(S_{new}\) by replacing \(w_j\)\;
                
                Compute \(new\_score = model.predict\_proba(S_{new})[Y]\)\;
                
                Add \((S_{new}, new\_score)\) to current candidates\;
            }
        }
    }

    Sort current candidates by score and keep top \(k\) in beam\;

    \ForEach{\((S_{cand}, score)\) in beam}{
        \If{model.predict(\(S_{cand}\)) \(\neq Y\)}{
            Add \((S_{cand}, score)\) to successful candidates\;
        }
    }

    \If{\# successful candidates \(\geq h\)}{
        \textbf{break}
    }
}

\BlankLine
Select \(S_{adv}\) from successful candidates using BLEURT similarity with original \(S\)\;

\KwRet \(S_{adv}\) if found, else None\;

\end{Algorithm}

%% file: hyper-parameters_vict_dataset.tex
\begin{table}[h!]
\caption{Selected hyper-parameter combinations for each combination of task and victim model.}
\label{table:hyper-parameters_task_victim}
\centering
\small
\begin{tabular}{llcccc}
\toprule
\textbf{} & \textbf{} & \textbf{Beam size} & \textbf{Hypotheses} & \textbf{Branches} & \textbf{Importances} \\
\textbf{Task} & \textbf{Victim} & \(k\) & \(h\) & \(b\) & \\
\midrule
PR2 & BiLSTM  & 25 & 30 & 50 & \texttt{logit-based} \\
 & BERT    & 60 & 10 & 80 & \texttt{logit-based} \\
 & RoBERTa & 60 & 10 & 80 & \texttt{logit-based} \\
\midrule
FC  & BiLSTM  & 10 & 10 & 30 & \texttt{logit-based} \\
  & BERT    & 10 & 10 & 30 & LIME \\
  & RoBERTa & 60 & 10 & 80 & \texttt{logit-based} \\
\midrule
RD  & BiLSTM  & 10 & 10 & 30 & \texttt{logit-based} \\
  & BERT    & 10 & 10 & 10 & \texttt{logit-based} \\
  & RoBERTa & 20 & 20 & 40 & \texttt{logit-based} \\
\midrule
HN  & BiLSTM  & 20 & 10 & 20 & \texttt{logit-based} \\
  & BERT    & 20 & 5  & 20 & LIME \\
  & RoBERTa & 10 & 10 & 10 & LIME \\
\midrule
C19 & BiLSTM  & 15 & 10 & 30 & \texttt{logit-based} \\
 & BERT    & 10 & 10 & 30 & \texttt{logit-based} \\
 & RoBERTa & 40 & 10 & 60 & \texttt{logit-based} \\
\bottomrule
\end{tabular}
\end{table}

%% file: running_time_vict_dataset.tex
\begin{table}[h!]
\caption{Running time statistics (in seconds) for each combination of task and victim model, measured on the entire dataset.}
\label{table:running_time_task_victim}
\centering
\small
\begin{tabular}{llrrr}
\toprule
 & & Time per & Total Attack & Total Evaluation \\
\textbf{Task} & \textbf{Victim} & Example (s) & Time (s) & Time (s) \\
\midrule
PR2 & BiLSTM  & 1.513   & 629.279    & 15.843 \\
 & BERT    & 12.586  & 5235.587   & 15.986 \\
 & RoBERTa & 72.192  & 30031.747  & 20.433 \\
\midrule
FC  & BiLSTM  & 2.111   & 854.859    & 54.206 \\
  & BERT    & 7.988   & 3235.195   & 51.033 \\
  & RoBERTa & 3.467   & 1403.967   & 44.047 \\
\midrule
RD  & BiLSTM  & 13.248  & 5497.992   & 560.019 \\
  & BERT    & 70.347  & 29193.831  & 488.945 \\
  & RoBERTa & 72.192  & 30031.747  & 20.433 \\
\midrule
HN  & BiLSTM  & 9.578   & 3831.309   & 615.956 \\
  & BERT    & 86.049  & 34419.532  & 660.498 \\
  & RoBERTa & 144.188 & 57675.011  & 544.444 \\
\midrule
C19 & BiLSTM  & 2.260   & 1344.562   & 95.838 \\
 & BERT    & 12.453  & 7409.627   & 103.194 \\
 & RoBERTa & 54.731  & 32564.714  & 117.173 \\
\bottomrule
\end{tabular}
\end{table}

%% file: num_queries.tex
\begin{table*}[h!]
\caption{Comparison of number queries on BiLSTM, BERT, and RoBERTa classifiers across five misinformation detection tasks. OpenFact did not publish their number of queries.}
\label{tab:num_queries}
\centering
\small
\begin{tabular}{llrrr}
\toprule
Task & Method & BiLSTM & BERT & RoBERTa \\
\midrule
\multirow{4}{*}{HN} 
& BERT-ATTACK & \cellcolor{gray!15}487.9 & \cellcolor{blue!15}648.4 & \cellcolor{green!15}1782.0 \\
& DeepWordBug & \cellcolor{gray!15}396.2 & \cellcolor{blue!15}395.9 & \cellcolor{green!15}384.3 \\
& BeamAttack & \cellcolor{gray!15}937.0 & \cellcolor{blue!15}4327.7 & \cellcolor{green!15}4596.6 \\
\midrule
\multirow{4}{*}{PR} 
& BERT-ATTACK & \cellcolor{gray!15}61.4 & \cellcolor{blue!15}80.2 & \cellcolor{green!15}117.6 \\
& DeepWordBug & \cellcolor{gray!15}27.5 & \cellcolor{blue!15}27.4 & \cellcolor{green!15}26.9 \\
& BeamAttack & \cellcolor{gray!15}593.4 & \cellcolor{blue!15}1373.4 & \cellcolor{green!15}10286.8 \\
\midrule
\multirow{4}{*}{FC}
& BERT-ATTACK & \cellcolor{gray!15}132.8 & \cellcolor{blue!15}146.7 & \cellcolor{green!15}164.1 \\
& DeepWordBug & \cellcolor{gray!15}54.4 & \cellcolor{blue!15}54.3 & \cellcolor{green!15}53.4 \\
& BeamAttack & \cellcolor{gray!15}1549.1 & \cellcolor{blue!15}1390.8 & \cellcolor{green!15}498.9 \\
\midrule
\multirow{4}{*}{RD}
& BERT-ATTACK & \cellcolor{gray!15}985.5 & \cellcolor{blue!15}774.3 & \cellcolor{green!15}951.9 \\
& DeepWordBug & \cellcolor{gray!15}232.8 & \cellcolor{blue!15}232.7 & \cellcolor{green!15}229.6 \\
& BeamAttack & \cellcolor{gray!15}3831.7 & \cellcolor{blue!15}10618.9 & \cellcolor{green!15}15458.1 \\
\midrule
\multirow{4}{*}{C19}
& BERT-ATTACK & \cellcolor{gray!15}127.2 & \cellcolor{blue!15}161.7 & \cellcolor{green!15}198.3 \\
& DeepWordBug & \cellcolor{gray!15}61.2 & \cellcolor{blue!15}61.0 & \cellcolor{green!15}61.0 \\
& BeamAttack & \cellcolor{gray!15}837.4 & \cellcolor{blue!15}2628.9 & \cellcolor{green!15}6491.4 \\
\bottomrule
\end{tabular}
\end{table*}

%% file: importance_methods.tex
\begin{table*}[ht!]
\caption{Evaluation metrics for different importance methods on different tasks and victims. We always bold the better BODEGA score with the same settings for either LIME or BERT. \( k \) refers to the selected beam size, \( h \) to hypothesis counts and \( b \) to the branching factor.}
\label{table:importance_methods}
\centering
\small
\begin{tabular}{llccrrrrr}
\toprule 
Task & Victim & \textbf{Importance Method} & \( k \), \( h \),  \( b \) & \makecell{Queries \\ per example} & Success & Semantic & Character & BODEGA \\
\midrule
 FC & BiLSTM & LIME & 10, 10, 30 & 2597.3 & 0.99 & 0.8111 & 0.9332 & 0.7511 \\
FC & BiLSTM & logit-based & 10, 10, 30 & 1549.1 & 1.00 & 0.8059 & 0.9410 & \textbf{0.7642} \\
\midrule
HN & BiLSTM & LIME & 10, 10, 20 & 1529.04 & 1.00 & 0.9043 & 0.9837 & 0.8905 \\
HN & BiLSTM & logit-based & 10, 10, 20 & 665.72 & 1.00 & 0.9112 & 0.9893 & \textbf{0.9031} \\
\midrule
\midrule
RD & BERT & LIME & 10, 10, 10 & 16342.9 & 0.70 & 0.8160 & 0.9174 & 0.5380 \\
RD & BERT & logit-based & 10, 10, 10 & 5947.0 & 0.90 & 0.8189 & 0.9130 & \textbf{0.6874} \\
\midrule
FC & BERT & LIME & 10, 10, 10 & 5520.44 & 1.00 & 0.8392 & 0.9546 & \textbf{0.8044} \\
FC & BERT & logit-based & 10, 10, 10 & 732.66 & 1.00 & 0.8188 & 0.9466 & 0.7787 \\
\midrule
\midrule
FC & RoBERTa & LIME & 10, 10, 20 & 557.08 & 1.00 & 0.8398 & 0.9644 & \textbf{0.8110} \\
FC & RoBERTa & logit-based & 10, 10, 20 & 112.4 & 1.00 & 0.8392 & 0.9641 & 0.8105 \\
\midrule
C19 & RoBERTa & LIME & 40, 10, 40 & 9714.4 & 1.00 & 0.5625 & 0.8602 & 0.4860 \\
C19 & RoBERTa & logit-based & 40, 10, 40 & 5505.0 & 1.00 & 0.6541 & 0.8939 & \textbf{0.5853} \\
C19 & RoBERTa & LIME & 10, 10, 20 & 1858.0 & 1.00 & 0.6304 & 0.8669 & \textbf{0.5500} \\
C19 & RoBERTa & logit-based & 10, 10, 20 & 972.7 & 1.00 & 0.6120 & 0.8677 & 0.5316 \\
\bottomrule
\end{tabular}
\end{table*}

%% file: scores_hypothesis_count.tex
\begin{table}[htbp!]
\caption{Results of different hypothesis counts \( h \) with the \textbf{RoBERTa} victim and different tasks.}
\label{tab:scores_hypothesis_count}
\centering
\begin{tabular}{lccccrrrrr}
\toprule
Task &  \textbf{\makecell{ hypothesis \\ count \( h \)}} & \makecell{beam \\ size \( k \)} & \makecell{branching \\ factor \( b \)} & Subset size & \makecell{Queries \\ per example} & Success & Semantic & Character & BODEGA \\
\midrule
RD  & 20 & 20 & 40 & 20& 37576.3 & 0.95 & 0.6645 & 0.7952 & 0.5501 \\
RD  & 10 & 20 & 40 & 20 & 37764.7 & 0.95 & 0.6654 & 0.7951 & 0.5508 \\
\midrule
PR2 & 20 & 40 & 40 & 30& 3744.8  & 1.00  & 0.5996 & 0.8235 & 0.5101 \\
PR2 & 10 & 40 & 40 & 30 & 3656.9  & 1.00  & 0.5993 & 0.8280 & 0.5130 \\
\midrule
C19 & 20 & 40 & 40 & 10 & 5505.0  & 1.00  & 0.6541 & 0.8939 & 0.5853 \\
C19 & 10 & 40 & 40 & 10 & 5505.0  & 1.00  & 0.6541 & 0.8939 & 0.5853 \\
\bottomrule
\end{tabular}
\end{table}

%% file: scores_beamsize.tex
\begin{table}[htbp!]
    \caption{Scores for Different Beam Sizes \( k \) for task \textbf{PR2} and victim \textbf{RoBERTa}. Here hypothesis counts \( h \) is always 10 and \( b \) = \( k \) for each case.}
    \label{tab:scores_beamsize}
    \centering
    \begin{tabular}{ccrrrrr}
        \toprule
        \makecell{\textbf{beam size \( k \)} \\ branching factor \( b \)} & Subset size & \makecell{Queries \\ per example} & Success & Semantic & Character & BODEGA \\
        \midrule
        10 & 30 & \textbf{530.53} & 0.93 & 0.58 & 0.78 & 0.44 \\
        20 & 30 & 1334.07 & 1.00 & 0.56 & 0.78 & 0.47 \\
        30 & 30 & 2451.77 & 1.00 & 0.58 & 0.79 & 0.49 \\
        35 & 30 & 3090.13 & 1.00 & 0.59 & 0.81 & 0.50 \\
        40 & 30 & 3656.93 & 1.00 & 0.60 & 0.83 & 0.51 \\
        50 & 30 & 5308.23 & 1.00 & 0.60 & \textbf{0.84} & 0.52 \\
        60 & 30 & 7151.87 & 1.00 & 0.61 & \textbf{0.84} & 0.53 \\
        70 & 30 & 9279.90 & 1.00 & 0.61 & 0.83 & 0.53 \\
        80 & 30 & 11374.77 & 1.00 & 0.62 & \textbf{0.84} & 0.54 \\
        90 & 30 & 14546.27 & 1.00 & 0.62 & \textbf{0.84} & 0.54 \\
        100 & 30 & 16499.97 & 1.00 & \textbf{0.63} & \textbf{0.84} & \textbf{0.55} \\
        \bottomrule
    \end{tabular}
\end{table}

%% file: scores_width.tex
\begin{table}[htbp!]
    \caption{Scores for Different branching factors \( b \) with a fixed Beam size \( k \) for task \textbf{PR2} and victim \textbf{RoBERTa} with hypothesis counts \( h \) = 10.}
    \label{tab:scores_width}
    \centering
    \begin{tabular}{cccrrrrr}
        \toprule
        \textbf{branching factor \( b \)} & Beam size  \( k \)& Subset size & \makecell{Queries \\ per example} & Success & Semantic & Character & BODEGA \\
        \midrule
        30 & 40 & 30 & \textbf{3032.733} & 1.0 & 0.5835 & 0.7924 & 0.4904 \\
        40 & 40 & 30 & 3656.933 & 1.0 & 0.5993 & 0.8280 & 0.5130 \\
        50 & 40 & 30 & 4476.300 & 1.0 & 0.5957 & \textbf{0.8331} & 0.5115 \\
        60 & 40 & 30 & 4836.333 & 1.0 & 0.6058 & 0.8309 & 0.5210 \\
        70 & 40 & 30 & 5799.533 & 1.0 & \textbf{0.6113} & 0.8249 & \textbf{0.5253} \\
        \bottomrule
    \end{tabular}
\end{table}

%% file: pos_mapping.tex
\begin{table}[H]
\caption{Mapping from Penn Treebank POS tags (as used by Flair) to Universal POS tags.}
\label{tab:pos_mapping}
\centering
\small
\begin{tabular}{lll}
\toprule
\textbf{Penn Tag} & \textbf{Description} & \textbf{Universal POS} \\
\midrule
ADD & Email address & X \\
AFX & Affix & X \\
CC & Coordinating conjunction & CCONJ \\
CD & Cardinal number & NUM \\
DT & Determiner & DET \\
EX & Existential there & PRON \\
FW & Foreign word & X \\
HYPH & Hyphen & PUNCT \\
IN & Preposition or subord. conjunction & ADP \\
JJ & Adjective & ADJ \\
JJR & Adjective, comparative & ADJ \\
JJS & Adjective, superlative & ADJ \\
LS & List item marker & X \\
MD & Modal & AUX \\
NFP & Superfluous punctuation & PUNCT \\
NN & Noun, singular or mass & NOUN \\
NNP & Proper noun, singular & PROPN \\
NNPS & Proper noun, plural & PROPN \\
NNS & Noun, plural & NOUN \\
PDT & Predeterminer & DET \\
POS & Possessive ending & PART \\
PRP & Personal pronoun & PRON \\
PRP\$ & Possessive pronoun & PRON \\
RB & Adverb & ADV \\
RBR & Adverb, comparative & ADV \\
RBS & Adverb, superlative & ADV \\
RP & Particle & PART \\
SYM & Symbol & SYM \\
TO & to & PART \\
UH & Interjection & INTJ \\
VB & Verb, base form & VERB \\
VBD & Verb, past tense & VERB \\
VBG & Verb, gerund or present participle & VERB \\
VBN & Verb, past participle & VERB \\
VBP & Verb, non-3rd person present & VERB \\
VBZ & Verb, 3rd person singular present & VERB \\
WDT & Wh-determiner & DET \\
WP & Wh-pronoun & PRON \\
WP\$ & Possessive wh-pronoun & PRON \\
WRB & Wh-adverb & ADV \\
XX & Unknown & X \\
-LRB- & Left round bracket & PUNCT \\
-RRB- & Right round bracket & PUNCT \\
. & Sentence-final punctuation & PUNCT \\
, & Comma & PUNCT \\
: & Colon or ellipsis & PUNCT \\
`` & Opening quotation mark & PUNCT \\
'' & Closing quotation mark & PUNCT \\
" & Quotation mark & PUNCT \\
\$ & Dollar sign & SYM \\
\bottomrule
\end{tabular}
\end{table}

%% file: main.bbl
\begin{thebibliography}{10}
\providecommand{\url}[1]{\texttt{#1}}
\providecommand{\urlprefix}{URL }
\providecommand{\doi}[1]{https://doi.org/#1}

\bibitem{akbik-etal-2019-flair}
Akbik, A., Bergmann, T., Blythe, D., Rasul, K., Schweter, S., Vollgraf, R.: {FLAIR}: An easy-to-use framework for state-of-the-art {NLP}. In: Ammar, W., Louis, A., Mostafazadeh, N. (eds.) Proceedings of the 2019 Conference of the North {A}merican Chapter of the Association for Computational Linguistics (Demonstrations). pp. 54--59. Association for Computational Linguistics, Minneapolis, Minnesota (Jun 2019). \doi{10.18653/v1/N19-4010}, \url{https://aclanthology.org/N19-4010/}

\bibitem{adversarial-text-genetic-optimization}
Alzantot, M., Sharma, Y., Elgohary, A., Ho, B.J., Srivastava, M., Chang, K.W.: Generating natural language adversarial examples. In: Riloff, E., Chiang, D., Hockenmaier, J., Tsujii, J. (eds.) Proceedings of the 2018 Conference on Empirical Methods in Natural Language Processing. pp. 2890--2896. Association for Computational Linguistics, Brussels, Belgium (Oct-Nov 2018). \doi{10.18653/v1/D18-1316}, \url{https://aclanthology.org/D18-1316}

\bibitem{clef-checkthat-2024}
Barr{\'o}n-Cede{\~{n}}o, A., Alam, F., Chakraborty, T., Elsayed, T., Nakov, P., Przyby{\l}a, P., Stru{\ss}, J.M., Haouari, F., Hasanain, M., Ruggeri, F., Song, X., Suwaileh, R.: The clef-2024 checkthat! lab: Check-worthiness, subjectivity, persuasion, roles, authorities, and adversarial robustness. In: Goharian, N., Tonellotto, N., He, Y., Lipani, A., McDonald, G., Macdonald, C., Ounis, I. (eds.) Advances in Information Retrieval. pp. 449--458. Springer Nature Switzerland, Cham (2024)

\bibitem{white-box-adversarial-attack-text-classification}
Ebrahimi, J., Rao, A., Lowd, D., Dou, D.: {H}ot{F}lip: White-box adversarial examples for text classification. In: Gurevych, I., Miyao, Y. (eds.) Proceedings of the 56th Annual Meeting of the Association for Computational Linguistics (Volume 2: Short Papers). pp. 31--36. Association for Computational Linguistics, Melbourne, Australia (Jul 2018). \doi{10.18653/v1/P18-2006}, \url{https://aclanthology.org/P18-2006}

\bibitem{replace-with-oov-attack}
Gao, J., Lanchantin, J., Soffa, M.L., Qi, Y.: Black-box generation of adversarial text sequences to evade deep learning classifiers. In: 2018 IEEE Security and Privacy Workshops (SPW). pp. 50--56 (2018). \doi{10.1109/SPW.2018.00016}

\bibitem{BAE}
Garg, S., Ramakrishnan, G.: {BAE}: {BERT}-based adversarial examples for text classification. In: Webber, B., Cohn, T., He, Y., Liu, Y. (eds.) Proceedings of the 2020 Conference on Empirical Methods in Natural Language Processing (EMNLP). pp. 6174--6181. Association for Computational Linguistics, Online (Nov 2020). \doi{10.18653/v1/2020.emnlp-main.498}, \url{https://aclanthology.org/2020.emnlp-main.498/}

\bibitem{iyyer2018adversarial}
Iyyer, M., Wieting, J., Gimpel, K., Zettlemoyer, L.: Adversarial example generation with syntactically controlled paraphrase networks. In: Walker, M., Ji, H., Stent, A. (eds.) Proceedings of the 2018 Conference of the North {A}merican Chapter of the Association for Computational Linguistics: Human Language Technologies, Volume 1 (Long Papers). pp. 1875--1885. Association for Computational Linguistics, New Orleans, Louisiana (Jun 2018). \doi{10.18653/v1/N18-1170}, \url{https://aclanthology.org/N18-1170/}

\bibitem{adversarial-attack-replace-with-grammatical-synonym}
Jin, D., Jin, Z., Zhou, J.T., Szolovits, P.: Is bert really robust? a strong baseline for natural language attack on text classification and entailment. Proceedings of the AAAI Conference on Artificial Intelligence  \textbf{34}(05),  8018--8025 (Apr 2020). \doi{10.1609/aaai.v34i05.6311}, \url{https://ojs.aaai.org/index.php/AAAI/article/view/6311}

\bibitem{levenshtein1966binary}
Levenshtein, V.I., et~al.: Binary codes capable of correcting deletions, insertions, and reversals. In: Soviet physics doklady. vol.~10, pp. 707--710. Soviet Union (1966)

\bibitem{lewoniewski2024openfact}
Lewoniewski, W., Stolarski, P., Str{\'o}{\.z}yna, M., Lewa{\'n}ska, E., Wojewoda, A., Ksi{\k{e}}{\.z}niak, E., Sawi{\'n}ski, M.: Openfact at checkthat! 2024: combining multiple attack methods for effective adversarial text generation. arXiv preprint arXiv:2409.02649  (2024), \url{https://arxiv.org/abs/2409.02649}

\bibitem{textbugger-word-char-lvl}
Li, J., Ji, S., Du, T., Li, B., Wang, T.: Textbugger: Generating adversarial text against real-world applications. In: Proceedings 2019 Network and Distributed System Security Symposium. NDSS 2019, Internet Society (2019). \doi{10.14722/ndss.2019.23138}, \url{http://dx.doi.org/10.14722/ndss.2019.23138}

\bibitem{bert-attack}
Li, L., Ma, R., Guo, Q., Xue, X., Qiu, X.: {BERT}-{ATTACK}: Adversarial attack against {BERT} using {BERT}. In: Webber, B., Cohn, T., He, Y., Liu, Y. (eds.) Proceedings of the 2020 Conference on Empirical Methods in Natural Language Processing (EMNLP). pp. 6193--6202. Association for Computational Linguistics, Online (Nov 2020). \doi{10.18653/v1/2020.emnlp-main.500}, \url{https://aclanthology.org/2020.emnlp-main.500}

\bibitem{roberta}
Liu, Y., Ott, M., Goyal, N., Du, J., Joshi, M., Chen, D., Levy, O., Lewis, M., Zettlemoyer, L., Stoyanov, V.: Roberta: A robustly optimized bert pretraining approach (2019), \url{https://arxiv.org/abs/1907.11692}

\bibitem{query-efficient-importances}
Maheshwary, R., Maheshwary, S., Pudi, V.: A strong baseline for query efficient attacks in a black box setting. In: Moens, M.F., Huang, X., Specia, L., Yih, S.W.t. (eds.) Proceedings of the 2021 Conference on Empirical Methods in Natural Language Processing. pp. 8396--8409. Association for Computational Linguistics, Online and Punta Cana, Dominican Republic (Nov 2021). \doi{10.18653/v1/2021.emnlp-main.661}, \url{https://aclanthology.org/2021.emnlp-main.661/}

\bibitem{piedrahita2024texttrojaners}
Piedrahita, D.G., Fazla, A., Krauter, L.: Texttrojaners at checkthat! 2024: Robustness of credibility assessment with adversarial examples through beamattack. In: Working Notes of CLEF 2024-Conference and Labs of the Evaluation Forum, CLEF (2024)

\bibitem{clef-checkthat:2024:task6}
Przyby{\l}a, P., Wu, B., Shvets, A., Mu, Y., Sheang, K.C., Song, X., Saggion, H.: Overview of the {CLEF}-2024 {CheckThat}! lab task 6 on robustness of credibility assessment with adversarial examples (incrediblae). In: Faggioli, G., Ferro, N., Galuščáková, P., {García Seco de Herrera}, A. (eds.) Working Notes of CLEF 2024 - Conference and Labs of the Evaluation Forum. CLEF~2024, Grenoble, France (2024)

\bibitem{bodega-paper}
Przybyła, P., Shvets, A., Saggion, H.: Verifying the robustness of automatic credibility assessment. Natural Language Processing p. 1–29 (Nov 2024). \doi{10.1017/nlp.2024.54}, \url{http://dx.doi.org/10.1017/nlp.2024.54}

\bibitem{adversarial-attack-replace-with-wordnet}
Ren, S., Deng, Y., He, K., Che, W.: Generating natural language adversarial examples through probability weighted word saliency. In: Korhonen, A., Traum, D., M{\`a}rquez, L. (eds.) Proceedings of the 57th Annual Meeting of the Association for Computational Linguistics. pp. 1085--1097. Association for Computational Linguistics, Florence, Italy (Jul 2019). \doi{10.18653/v1/P19-1103}, \url{https://aclanthology.org/P19-1103}

\bibitem{lime}
Ribeiro, M.T., Singh, S., Guestrin, C.: "why should {I} trust you?": Explaining the predictions of any classifier. In: Proceedings of the 22nd {ACM} {SIGKDD} International Conference on Knowledge Discovery and Data Mining, San Francisco, CA, USA, August 13-17, 2016. pp. 1135--1144 (2016)

\bibitem{ribeiro-etal-2018-semantically}
Ribeiro, M.T., Singh, S., Guestrin, C.: Semantically equivalent adversarial rules for debugging {NLP} models. In: Gurevych, I., Miyao, Y. (eds.) Proceedings of the 56th Annual Meeting of the Association for Computational Linguistics (Volume 1: Long Papers). pp. 856--865. Association for Computational Linguistics, Melbourne, Australia (Jul 2018). \doi{10.18653/v1/P18-1079}, \url{https://aclanthology.org/P18-1079}

\bibitem{bleurt}
Sellam, T., Das, D., Parikh, A.: {BLEURT}: Learning robust metrics for text generation. In: Jurafsky, D., Chai, J., Schluter, N., Tetreault, J. (eds.) Proceedings of the 58th Annual Meeting of the Association for Computational Linguistics. pp. 7881--7892. Association for Computational Linguistics, Online (Jul 2020). \doi{10.18653/v1/2020.acl-main.704}, \url{https://aclanthology.org/2020.acl-main.704}

\bibitem{advAttacK_Survey}
Zhang, W.E., Sheng, Q.Z., Alhazmi, A., Li, C.: Adversarial attacks on deep-learning models in natural language processing: A survey. ACM Trans. Intell. Syst. Technol.  \textbf{11}(3) (apr 2020). \doi{10.1145/3374217}, \url{https://doi.org/10.1145/3374217}

\bibitem{10.1007/978-3-031-33377-4_35}
Zhu, H., Zhao, Q., Wu, Y.: Beamattack: Generating high-quality textual adversarial examples through beam search and mixed semantic spaces. In: Kashima, H., Ide, T., Peng, W.C. (eds.) Advances in Knowledge Discovery and Data Mining. pp. 454--465. Springer Nature Switzerland, Cham (2023)

\end{thebibliography}
